%% file: refactor.tex
\pdfoutput=1

\documentclass[11pt]{article}
\usepackage[most]{tcolorbox}
\usepackage[dvipsnames]{xcolor}

\usepackage[preprint]{acl}
\usepackage{xcolor}
\usepackage{booktabs}

\usepackage{times}
\usepackage{latexsym}
\usepackage{float}
\usepackage{enumitem}
\usepackage{url}
\usepackage{tcolorbox}
\usepackage{listings}
\tcbuselibrary{listingsutf8}
\usepackage{xcolor}

\usepackage[T1]{fontenc}

\usepackage[utf8]{inputenc}

\usepackage{booktabs}  
\usepackage{array}     

\usepackage{microtype}

\usepackage{inconsolata}
\usepackage[absolute,overlay]{textpos}
\usepackage{graphicx}

\usepackage{graphicx}

\newcommand{\gptfourone}{GPT-4.1}

\title{Beyond “Not Novel Enough”: Enriching Scholarly Critique with LLM-Assisted Feedback}

\author{
 \textbf{Osama Mohammed Afzal\textsuperscript{1}},
 \textbf{Preslav Nakov\textsuperscript{2}},
 \textbf{Tom Hope\textsuperscript{3}},
 \textbf{Iryna Gurevych\textsuperscript{1}}
\\\\
 \textsuperscript{1} UKP Lab, TU Darmstadt and Hessian Center for AI (hessian.AI) \\ 
 \textsuperscript{2}MBZUAI,
 \textsuperscript{3}The Allen Institute for AI (AI2),
\vspace{0.25em}\\
 \url{www.ukp.tu-darmstadt.de}\\
}

\begin{document}
\maketitle

\begin{abstract}
Novelty assessment is a central yet understudied aspect of peer review, particularly in high-volume fields like NLP where reviewer capacity is increasingly strained. We present a structured approach for automated novelty evaluation that models expert reviewer behavior through three stages: content extraction from submissions, retrieval and synthesis of related work, and structured comparison for evidence-based assessment. Our method is informed by analysis of human-written novelty reviews and captures key patterns such as independent claim verification and contextual reasoning. Evaluated on 182 ICLR 2025 submissions with human annotated reviewer novelty assessments, the approach achieves 86.5\% alignment with human reasoning and 75.3\% agreement on novelty conclusions—substantially outperforming existing LLM-based baselines. The method produces detailed, literature-aware analyses and improves consistency over ad hoc reviewer judgments. These results highlight the potential for structured LLM-assisted approaches to support more rigorous and transparent peer review without displacing human expertise. Data and
code are made available. \footnote{\url{https://ukplab.github.io/eacl2026-assessing-paper-novelty/}}
\end{abstract}

\section{Introduction}


The peer review system is collapsing under its own success. Two independent committees at NeurIPS 2021 disagreed on 23\% of identical papers~\cite{beygelzimer2023machinelearningreviewprocess}—a breakdown in consistency that signals deeper problems than mere capacity constraints. With manuscript submissions doubling every 15 years~\cite{RePEc:spr:scient:v:84:y:2010:i:3:d:10.1007_s11192-010-0202-z} and reviewers now handling 14 evaluations annually~\cite{Diaz2024Annotations}, the system's 15 million annual reviewing hours~\cite{Aczel2021} are producing increasingly unreliable outcomes.

Among peer review tasks, novelty assessment stands out as one of the most problematic~\cite{Ernst2020UnderstandingPR} \cite{Horbach2018TheAO}. Novelty assessment requires reviewers to determine whether a submission makes sufficiently original contributions by identifying what specific advances it makes beyond existing work, evaluating whether these advances are significant enough to warrant publication, and verifying that the authors have accurately characterized their contributions relative to prior research. This knowledge-intensive process demands that reviewers maintain comprehensive awareness of related work across their field and can precisely distinguish between meaningful innovations and incremental modifications—a task that becomes exponentially more difficult as publication rates accelerate and research domains specialize. Overwhelmed reviewers often resort to superficial analyses, producing vague feedback like "not novel enough" without clear justification. The challenge compounds when reviewers encounter papers outside their specific expertise, leading to either overly conservative rejections or inadequate assessments that fail to catch incremental work~\cite{kuznetsov2024naturallanguageprocessingpeer}.

Recent advances in large language models present an unprecedented opportunity to address these novelty assessment challenges at scale. These breakthrough technologies have revolutionized text processing and demonstrated remarkable performance across knowledge-intensive tasks~\cite{Raiaan2024ARO}, with recent technical advancements expanding capabilities to specialized reasoning and efficient inference~\cite{Li2024LLMIS, Zhang2025KAGThinkerIT}.

While recent LLM advances create this opportunity, no existing work specifically addresses novelty assessment as a dedicated task within the peer review process. Prior research incorporates novelty evaluation within idea generation pipelines~\cite{radensky2025scideatorhumanllmscientificidea,lu2024aiscientist,li2024chainideasrevolutionizingresearch}, generates peer reviews with novelty assessments occurring as a result of them existing in peer reviews from training data~\cite{idahl-ahmadi-2025-openreviewer,darcy2024margmultiagentreviewgeneration}, or adds novelty assessment steps to review synthesis pipelines for improvement~\cite{zhu2025deepreviewimprovingllmbasedpaper}. However, these approaches either operate on synthetic ideas rather than real research contributions or fail to evaluate novelty assessment capabilities in isolation. This represents a critical gap requiring specialized methodologies for peer review novelty assessment.

To address this gap, we propose an end-to-end novelty assessment pipeline for peer review submissions. Our approach consists of three stages: document processing and content extraction, related work retrieval and ranking, and structured novelty assessment. The final stage implements four sequential steps: novelty related content selection from the submission pdf, building comprehensive understanding of related work from retrieved papers, comparing claimed novelty against the comprehensive analysis from the prior step, and generating a summary with cited evidence from the comparison. This pipeline operates on real research papers and directly evaluates novelty assessment capabilities, addressing the limitations of existing approaches. Importantly, we conduct the first evaluation of LLMs for novelty assessment using actual human data, including annotated novelty assessment statements, and provide comprehensive evaluation across multiple dimensions.
\paragraph{Research Questions and Contributions}

This work aims to address the following research questions:

\begin{enumerate}
    \item How does our human-informed novelty assessment pipeline compare to existing approaches?
    \item How well do our assessments align with human reviewer preferences across key evaluation dimensions?
    \item Can automated evaluation reliably substitute for human judgment in assessing novelty assessment quality?
\end{enumerate}

Our contributions are threefold:

\begin{itemize}
    \item \textbf{Human Analysis Dataset and Insights}: A systematically curated dataset of 182 papers with annotated human novelty assessments from ICLR 2025, along with empirical insights into expert reviewer reasoning patterns, evaluation criteria, and argument structures that inform AI system design for novelty assessment.

    \item \textbf{Human-Informed Pipeline}: A literature-grounded pipeline that incorporates insights from human novelty assessment practices, featuring structured prompting strategies and targeted content extraction informed by observed expert reviewer behavior.

    \item \textbf{Comprehensive Evaluation and Analysis}: Systematic comparison of our human-informed approach against existing baselines and human reviewers, with fine-grained evaluation across multiple dimensions and validation of automated assessment methods.

\end{itemize}

\begin{figure*}[htbp]
    \centering
    \includegraphics[width=\textwidth]{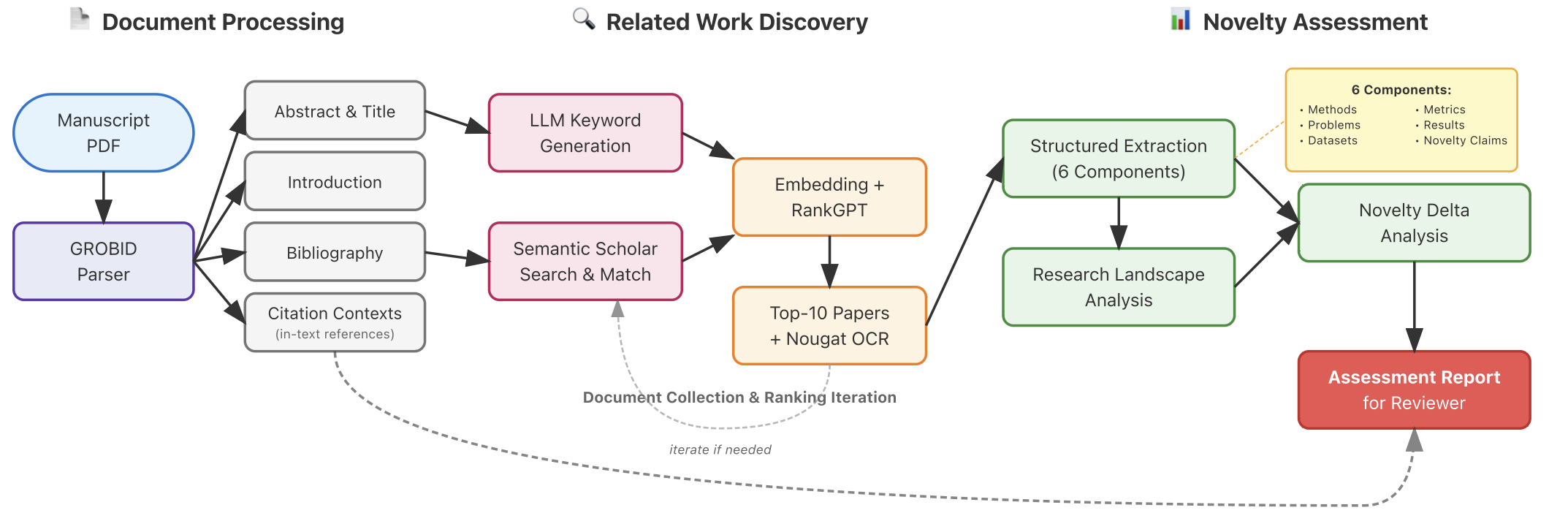}
    \caption{Automated novelty assessment pipeline. The system processes manuscripts through three stages: (1) Document Processing extracts content using GROBID, (2) Related Work Discovery identifies and ranks relevant papers via embedding similarity and LLM reranking, and (3) Novelty Assessment performs structured analysis to generate evidence-based novelty evaluations.}
    \label{fig:review-workflow}
\end{figure*}

\section{Related Work}
\paragraph{AI-Assisted Peer Review Systems}
Our work is positioned at the peer review stage of scientific research, where our system operates when a manuscript is submitted for evaluation. While previous works \cite{darcy2024margmultiagentreviewgeneration} \cite{idahl-ahmadi-2025-openreviewer} \cite{zhu2025deepreviewimprovingllmbasedpaper} \cite{chitale2025autorev} \cite{Chang2025TreeReviewAD} \cite{nemecek2025feasibilitytopicbasedwatermarkingacademic} have developed end-to-end peer review generation pipelines that may implicitly include novelty assessment steps, we are the first to focus specifically on building a dedicated pipeline for novelty assessment and the first to systematically evaluate LLMs on this task. A related line of work operates at the ideation stage of research \cite{radensky2025scideatorhumanllmscientificidea} \cite{shahid2025literature} \cite{li2024chainideasrevolutionizingresearch} \cite{lu2024aiscientist}, developing pipelines for research idea generation that aim to improve novelty through feedback loops from a novelty assessor. In contrast, we operate at a more mature stage where ideas have been fully executed and comparative analyses are well-formulated. The evaluation in ideation-stage works focuses on synthetic ideas that are typically abstract and loosely defined, whereas we evaluate concrete, polished research contributions that have undergone the refinement process of execution and manuscript preparation.

\paragraph{Scientific Literature Analysis \& Retrieval}
Our work employs an extensive related work discovery pipeline that collects papers cited within the submission and additionally retrieves related papers by querying with prompts generated by \gptfourone{}. Papers are then ranked using an embedding-based method and reranked using RankGPT. We adapt this general approach from existing work \cite{radensky2025scideatorhumanllmscientificidea}\cite{shahid2025literature}\cite{li2024chainideasrevolutionizingresearch} with modifications to ranking and filtering for our specific task. Similar retrieval-rank-rerank pipelines have been used for related work generation \cite{agarwal2025litllmtoolkitscientificliterature}. Another retrieval approach is OpenScholar \cite{asai2024openscholarsynthesizingscientificliterature}, which uses an LLM-RAG based approach to answer scientific queries by identifying relevant passages from 45 million open-access papers. Works like DeepReviewer \cite{zhu2025deepreviewimprovingllmbasedpaper} incorporate OpenScholar for novelty validation. However, our primary criticism of OpenScholar for novelty assessment is that it provides only generic comparisons rather than the granular analysis across methodology, problem formulation, evaluation approaches, and novelty claims that our task requires.

\paragraph{Evaluation of LLM Generated Text}
Prior works evaluating generated peer reviews have adopted either quantitative evaluations, where they compare LLM-assigned scores (such as Overall Score, Soundness, etc.) against human-assigned scores on review forms, or qualitative evaluations using traditional metrics like BERTScore \cite{zhang2020bertscoreevaluatingtextgeneration}, ROUGE \cite{lin-2004-rouge}, and BLEU \cite{papineni-etal-2002-bleu}, or more recent approaches like LLM-as-Judge \cite{zheng2023judging}. We adopt the LLM-as-Judge approach for our evaluation. Notably, no prior work has specifically evaluated LLM performance on novelty assessment as a dedicated task, making our evaluation framework the first of its kind.

\textcolor{black}{\textbf{Dataset Scale}: Our dataset comprises 182 papers and 352 reviews, which is comparable to or larger than datasets used in related peer review research: \cite{du-etal-2024-llms} use 100 papers with 380 reviews, \cite{kennard-etal-2022-disapere} label review data sourced for 188 papers, \cite{hua-etal-2019-argument} work with 400 reviews, and \cite{chamoun-etal-2024-automated} evaluate on 300 reviews. Novelty assessment requires careful manual annotation of scattered novelty discussions across reviews, making large-scale annotation resource-intensive. Following established practice in peer review analysis, we prioritize annotation quality over quantity, ensuring each example receives thorough annotation.}

\section{Methodology}
\subsection{Human Analysis for Prompt Design}
\label{sec:3.1}
To understand human novelty assessment, we analyzed ICLR 2025 reviews, which explicitly require novelty evaluation in dedicated sections. We sourced submissions from OpenReview and used keyword-based search for terms including "novel", "original", "research gap", "innovation", "incremental", "prior work", and "existing work". Papers were ranked by: (1) reviews with >4 novelty keywords, (2) consistent novelty discussion patterns, and (3) total review count. We selected the top 200 papers.
To expedite annotation, we used GPT-4o mini for sentence-level classification to identify novelty discussions, which a human annotator then verified and refined by selecting all sentences containing actual novelty assessments. This annotation process revealed that 18 papers (9\%) contained limited genuine novelty assessments—often keyword matches referring to paper components rather than evaluation. The remaining 182 papers formed our analysis dataset. We identified recurring patterns in reviewer reasoning, evaluation criteria, and argument structures, focusing on how reviewers structure arguments, prioritize evidence, and compare submissions to prior work.
\textcolor{black}{The four key patterns were identified through an exploratory qualitative review, where the primary author examined novelty-related review segments, allowing patterns to emerge inductively through close reading and thematic analysis.} This analysis revealed several key patterns in how expert reviewers assess novelty:

\textbf{Verification over acceptance:} Rather than accepting author claims at face value, reviewers independently verify relationships with prior work and critically examine how authors characterize related research, often distinguishing between author framing and actual technical relationships. Our prompt explicitly instructs models to "independently verify relationships" and "distinguish between author-claimed differences and independently observed differences," mirroring this critical verification approach, as shown in Figures~\ref{fig:novelty-prompt-1} and~\ref{fig:novelty-prompt-2}.

\textbf{Variable granularity:} Reviewers assess contributions with varying detail—some providing global novelty assessments while others examine each contribution separately against relevant prior work. (We address this through the "Contribution Delta Analysis" section that systematically examines each claimed contribution individually against the most similar prior work, ensuring comprehensive coverage regardless of author presentation style, as detailed in Figure~\ref{fig:novelty-prompt-2}.)

\textbf{Different analytical lenses:} Some reviewers focus on methodological innovations while others evaluate systems holistically, calibrating expectations based on field maturity. Our prompt incorporates multiple analytical perspectives through separate sections for research positioning, methodological relationships, and field context considerations that help calibrate novelty expectations based on area maturity, shown across Figures~\ref{fig:novelty-prompt-1} and~\ref{fig:novelty-prompt-2}.

\textbf{Gap identification:} Reviewers systematically identify gaps in related work discussions and distinguish between implementation-level improvements and genuine conceptual advances. (The "Related Work Considerations" section specifically instructs models to identify missing comparisons and assess whether improvements stem from "implementation details rather than conceptual advances," directly addressing this reviewer behavior in Figure~\ref{fig:novelty-prompt-2}.)
These insights informed both our prompt task design and the input to the LLM.

\subsection{Our Approach}

\paragraph{Overview}
Our pipeline processes submission PDFs and generates structured novelty assessments through three stages (Figure \ref{fig:review-workflow}): (i) Document Processing extracts key content from submissions, (ii) Related Work Discovery identifies and ranks relevant prior work, and (iii) Novelty Assessment performs comparative analysis to generate evidence-based novelty evaluations.

\subsection{Stage 1: Document Processing}
We extract structured content from submission PDFs using GROBID\footnote{https://github.com/kermitt2/grobid} to obtain titles, abstracts, bibliographies, and citation contexts required for subsequent stages.

\subsection{Stage 2: Related Work Discovery}
This stage identifies and ranks related work through a multi-step retrieval pipeline designed to capture both explicitly cited works and potentially relevant uncited research.

\paragraph{Cited Work Processing}
Bibliography entries are matched against Semantic Scholar to obtain standardized metadata (title, abstract, authors, publication date, venue) for consistent downstream processing.

\paragraph{Uncited Work Discovery}
To identify relevant work not cited by authors, we generate 5 keyword queries using GPT-4.1 and search Semantic Scholar. Results are filtered to remove exact title matches with the submission (avoiding potential preprints) and papers published after the submission date.

\paragraph{Embedding-based Ranking}
We generate embeddings for all collected papers using SPECTER v2~\cite{Singh2022SciRepEvalAM} on concatenated titles and abstracts. Papers are ranked by cosine similarity to the submission's embedding to identify semantically similar work.

\paragraph{LLM-based Reranking}
To prioritize papers with conceptual rather than purely semantic similarity, we employ LLM-based reranking~\cite{sun-etal-2023-chatgpt,Sun2023InstructionDM} with prompts emphasizing methodological approaches, novelty claims, and problem statements. We select the top-K (k=20) papers for novelty assessment.

\paragraph{Content Extraction}
For selected papers, we retrieve PDFs through a hierarchical search across Semantic Scholar, ACL Anthology, and arXiv. Retrieved papers are processed using MinerU~\cite{wang2024mineruopensourcesolutionprecise,he2024opendatalab} to extract introduction sections, with Nougat OCR~\cite{blecher2023nougat} as fallback for processing failures. We use these tools for OCRs here as they output more accurate OCRs and we will be using this paper content in the next stage.

\subsection{Stage 3: Novelty Assessment}

We use GPT-4.1~\cite{openai2024gpt41} for its improved instruction-following capabilities. This stage consists of four sequential steps.

\paragraph{Structured Extraction}
Processing retrieved papers as raw text creates context optimization challenges that degrade LLM performance. Recent research demonstrates that model performance consistently degrades with increasing input length, even when task complexity remains constant~\cite{hong2025context}. This occurs because either overwhelming models with unrelated information reduces accuracy~\cite{zhu2025deepreviewimprovingllmbasedpaper,idahl-ahmadi-2025-openreviewer} or insufficient context through heavy truncation limits understanding~\cite{radensky2025scideatorhumanllmscientificidea}.

We extract six structured components aligned with novelty assessment requirements from each paper's title, abstract, introduction: (i) Methods, (ii) Problems addressed, (iii) Datasets, (iv) Results, (v) Evaluation approaches, and (vi) Novelty Claims. This preserves essential information while reducing context length to mitigate the performance degradation observed with longer, unstructured inputs (Figure~\ref{fig:extraction-prompt}).

\textbf{Landscape Analysis} Expert reviewers possess comprehensive domain knowledge of established benchmarks, techniques, metrics, and recent developments in their areas. To approximate this foundation, we incorporate a landscape analysis step that systematically organizes structured components from retrieved related work.
Using GPT-4.1, we perform cross-paper synthesis to identify methodological clusters, trace problem evolution, map evaluation ecosystems, and establish technical relationships (Figure \ref{fig:landscape-prompt}). This produces a hierarchical organization of the research space with explicit connections between related, competing, and complementary approaches, providing contextual background for novelty assessment that mimics expert reviewers' organized domain understanding.

\paragraph{Novelty Delta Analysis}
This step performs comparative analysis between the submission and prior work using three inputs: (1) the research landscape, (2) the submission's claimed contributions, and (3) citation contexts—sentences where the submission cites related work. Citation contexts reveal how authors position their contributions, enabling verification of claimed distinctions versus rhetorical framing.
Using GPT-4.1 with prompts informed by our human analysis (Section \ref{sec:3.1}), the system implements key reviewer patterns: independent verification of author claims, variable granularity examination of contributions, and identification of gaps in related work discussions (Figures \ref{fig:novelty-prompt-1} and \ref{fig:novelty-prompt-2}).
\paragraph{Assessment Report Generation}
The final step generates a concise paragraph long summary that appears similar to actual peer review novelty assessments, enabling direct comparison with human-written assessments (Figure \ref{fig:summary-prompt}).

\section{Evaluation}

\input{tables/dataset_count}

\input{tables/JudgementEval}

\subsection{Evaluation Data}
The evaluation dataset comprises the same 182 annotated examples used during human prompt design. For each example, we prompt \gptfourone{} with the human review and its corresponding annotated novelty statements to generate a coherent assessment using the prompt in Figure \ref{fig:normalization-prompt}. This synthesis step is necessary because novelty-related comments in reviews are typically scattered rather than consolidated. Direct concatenation of these fragments would introduce stylistic biases during evaluation, as the fragmented human annotations would differ substantially in structure and coherence from the unified assessments generated by our system. We therefore use the \gptfourone{}-synthesized assessments as our ground truth in evaluation. To assess the risk of potential data leakage into GPT-4.1’s pre-training corpus, we examined our evaluation set of 182 ICLR 2025 submissions. Only 11 of these papers had appeared on arXiv prior to the model’s knowledge cutoff of June 1, 2024.

\subsection{Evaluation Methods}
\paragraph{Automated Evaluation}
Evaluating novelty assessment systems presents significant challenges due to the subjective and knowledge-intensive nature of the task. What constitutes "novel" depends heavily on the evaluator's familiarity with the surrounding research landscape. Even when human reviewers reach similar novelty conclusions, they may arrive at these decisions through different reasoning paths and evidence bases. 


Given these challenges, we employ an LLM-as-Judge framework using our style-normalized human novelty assessments as ground truth. We evaluate AI-generated assessments across four key dimensions using the prompts in Figures \ref{fig:prompt-extraction-core-judgments} and \ref{fig:prompt-reviewer-evaluation} with \gptfourone{} as our Judge:

\textbf{Novelty Conclusion Alignment}: Whether the AI assessment reaches similar novelty conclusions as human reviewers.

\textbf{Novelty Reasoning Alignment}: Whether the AI's reasoning process and justifications align with human reviewer logic.

\textbf{Prior Work Engagement}: Whether the assessment demonstrates adequate engagement with relevant literature rather than superficial analysis.

\textbf{Depth of Analysis}: Whether the assessment provides substantive, detailed evaluation rather than surface-level observations.

These dimensions ensure that AI assessments not only align with human judgments but also meet quality standards for thorough, evidence-based novelty evaluation. Our evaluation employs a two-stage process to ensure consistency. First, we extract core judgments (key novelty strengths and weaknesses) from human reviews using \gptfourone{} with the prompt in Figure \ref{fig:prompt-extraction-core-judgments}. We perform this extraction separately to establish stable reference judgments, as combining extraction with evaluation would risk the LLM identifying different claims across comparisons. In the second stage, we evaluate AI-generated assessments against these pre-extracted judgments using the prompt in Figure \ref{fig:prompt-reviewer-evaluation}. This evaluation quantifies four aspects: (1) \textit{judgment similarity}, measuring whether the AI identifies the same specific novelty aspects with confidence scores; (2) \textit{conclusion alignment}, checking whether bottom-line novelty sufficiency verdicts match; (3) \textit{prior work engagement}, categorized as None, Limited (1-2 citations), or Extensive (3+); and (4) \textit{depth of analysis}, rated as Surface Level, Moderate (1-2 aspects), or Deep (3+ detailed comparisons). Table \ref{tab:overall_summary} reports the resulting alignment scores across these dimensions.

\paragraph{Human Evaluation}
We validate our automated evaluation with three PhD students (two third-year, one first-year) in NLP and AI for Science, all with multiple publications. They perform pairwise comparisons across the same four dimensions, viewing side-by-side novelty assessments from humans, our system, and baselines. We collect 100 total judgments: 25 shared samples per annotator (for agreement) and 25 unique samples each, sampled randomly.
For each comparison, annotators choose A, B, Tie, or Unclear and may optionally provide comments (Figures~\ref{fig:human-eval-ui}, \ref{fig:human-eval-ui-2}). Table~\ref{tab:irr_metrics} reports moderate inter-rater agreement (0.493–0.560) and fair kappa scores (0.287–0.368), consistent with the subjectivity of novelty evaluation.

\subsection{Baseline Methods}
We compare our approach against three existing systems, adapting each for novelty assessment evaluation.

\paragraph{Scideator \cite{radensky2025scideatorhumanllmscientificidea}}
Scideator includes a novelty classification module that uses GPT-4o with few-shot examples and task definition to classify ideas as 'novel' or 'not novel'. Originally designed for iterative idea refinement, we adapt it by using paper titles and abstracts as input instead of nascent ideas.

\paragraph{OpenReviewer \cite{idahl-ahmadi-2025-openreviewer}}
OpenReviewer generates comprehensive peer reviews using Llama-OpenReviewer-8B, trained on 79,000 expert reviews from top conferences. We extract novelty-related content from its outputs using the same LLM-based approach applied to human reviews (Figure \ref{fig:normalization-prompt}).

\paragraph{DeepReviewer \cite{zhu2025deepreviewimprovingllmbasedpaper}}
DeepReviewer is a multi-stage review framework that combines literature retrieval done with OpenScholar \cite{asai2024openscholarsynthesizingscientificliterature} with evidence-based argumentation, powered by DeepReviewer-14B trained on structured review annotations. We extract novelty assessments using the same approach as OpenReviewer. Notably, DeepReviewer was trained on ICLR 2025 data, which includes our \textbf{entire evaluation} dataset.

\paragraph{Adaptation of Baselines}
Our groundtruth novelty assessments are extracted from human written reviews from the ICLR 2025 set that we labeled earlier. These extracted novelty segments are then run via the style normalization prompt in figure \ref{fig:normalization-prompt}. Similarly both the DeepReviewer and OpenReviewer produced peer reviews are run via this pipeline to extract novelty segments and compose a coherent novelty assessment and run through the style normalization module. This is a fair adaptation as given these models are trained on these peer review datasets and they should able to mimic the distribution of novelty assessments found in these peer reviews. Given deepreviewer is trained on the data we are evaluating so even reproducing their training data would be enough to score high on the evaluation set.

\section{Results and Analysis}

We evaluated each system by comparing its novelty assessments against human novelty assessments as reference. For papers with multiple human reviewers, we also conducted human-vs-human comparisons to establish a baseline. Table \ref{tab:overall_summary} presents the overall results.

\subsection{Overall Performance}

\begin{figure}[t]
\centering
\includegraphics[width=0.95\columnwidth]{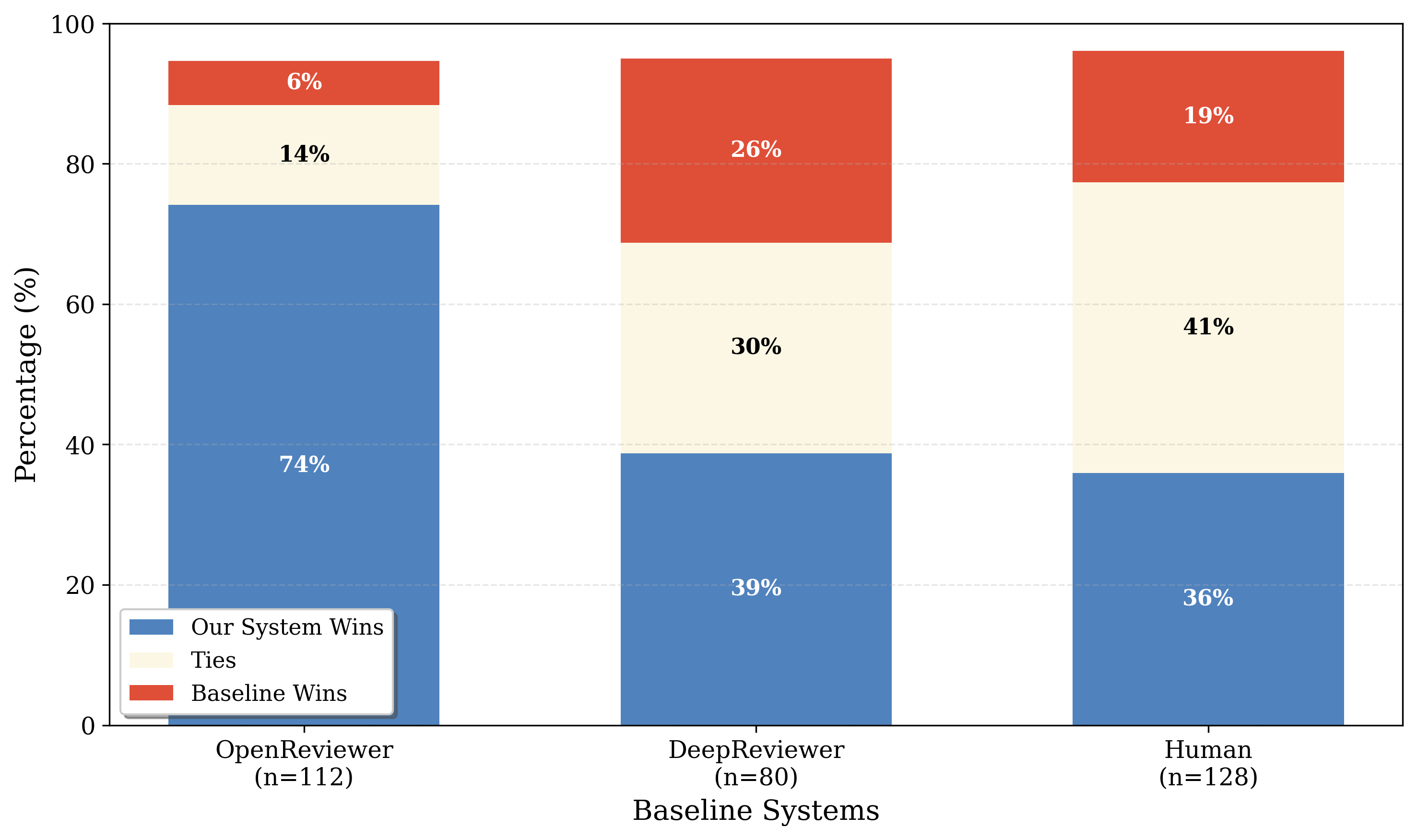}
\caption{Overall performance comparison between our system and three baseline systems based on human evaluation (n values indicates number of comparisons)}
\label{fig:human_eval_sys}
\end{figure}

Our system significantly outperforms both AI baselines and the human-vs-human baseline. For Reasoning Alignment, we achieve scores 44.1 and 35.9 percentage points higher than OpenReviewer \cite{idahl-ahmadi-2025-openreviewer} and DeepReviewer \cite{zhu2025deepreviewimprovingllmbasedpaper}, and 21.4 points above the human baseline. For Conclusion Agreement, our system again leads all baselines, outperforming the nearest human baseline by approximately 13 percentage points (Tables \ref{tab:full-text-comparison}, \ref{tab:durnd-comparison}, \ref{tab:meta-instructions-comparison}).

\paragraph{Sentiment Shift Analysis}
We analyze Positive Shift (neutral/negative → positive sentiment vs. human reference) and Negative Shift (the opposite). AI systems show optimistic bias, with DeepReviewer exhibiting high Positive Shift. Our system shows lower Positive Shift than DeepReviewer, though OpenReviewer aligns most closely with human rates. For Negative Shift, OpenReviewer mirrors humans' critical tendency, followed by DeepReviewer. Our approach achieves the lowest Negative Shift rate.

\paragraph{Depth and Prior Work Engagement}
Our system achieves the highest scores for both dimensions, producing no surface-level analyses unlike all baselines (Tables \ref{tab:depth} and \ref{tab:priorwork}). This stems from our specialized multi-step pipeline targeting novelty assessment, while other systems generate complete peer reviews where novelty is secondary. OpenReviewer performs worst, lacking retrieval capabilities. DeepReviewer uses OpenScholar retrieval but fails at comparative analysis. Human reviewers show high variance in engagement depth.

\paragraph{Human Evaluation Validation}

\input{tables/ablations}
Human evaluations validate our LLM-as-Judge framework. Our system wins 74\% of comparisons against OpenReviewer (Figures \ref{fig:human_eval_sys} and \ref{fig:human_eval_dim}). Against DeepReviewer and human reviewers, win rates are lower (39\% and 36\%), but high tie rates (30\% and 41\%) indicate comparable quality, with low loss rates (16-26\%). By dimension, Claim Substantiation and Analytical Quality achieve the highest win rates (56\% and 55\%), while Novelty Decision shows the most ties (31\%), suggesting similar conclusions across approaches. These patterns align with automated results, supporting our evaluation framework's validity.

\subsection{Analysis: Understanding Human Alignment Patterns}
\label{sec:human-alignment}

Our system's higher agreement scores compared to human-human baselines warrant careful examination. To investigate this, we analyzed papers with multiple human reviewers to understand the sources of disagreement. \textcolor{black}{We detail the analysis methodology in Appendix \ref{sec:analysis}.}

\paragraph{Sources of Human Reviewer Variability}

Qualitative analysis reveals several factors contributing to reviewer disagreement:
\textbf{Different Evaluation Lenses}: Reviewers often focus on different aspects of novelty. In submission Ipe4fMCBXk, half the reviewers emphasized methodological contributions while others focused on application novelty, leading to opposite conclusions from the same paper. \textbf{Varying Domain Expertise}: Reviewers' background knowledge affects assessments. For instance, in a protein design paper, reviewers familiar with the field's history correctly identified prior work on recombination techniques, while others assessed these as novel contributions. \textbf{Assessment Granularity}: Some reviewers provide high-level judgments ("innovative approach") while others focus on specific technical details. This variation in granularity contributes to disagreement even when reviewers might agree on underlying facts.

\paragraph{The Role of Systematic Evaluation}
Our system's approach differs from human review in applying consistent evaluation criteria. It evaluates multiple dimensions (methodology, application, prior work) for every paper, maintains uniform depth of analysis across assessments, and applies consistent thresholds for novelty judgments. This systematic approach may explain the alignment patterns: when human reviewers disagree due to focusing on different aspects, our system's comprehensive evaluation can align partially with each perspective.

\subsection{Component Analysis}
Table \ref{tab:component_analysis} shows the incremental contribution of each pipeline component. Our human-informed prompt design provides the largest gains (+40.7\% reasoning, +46.8\% conclusion), reflecting the importance of structured evaluation criteria derived from our human analysis. Structured extraction adds moderate improvements (+3.3\% reasoning, +4.5\% conclusion) but reduces overall computation costs and time, while landscape analysis contributes minimally (+3.2\% reasoning, -0.7\% conclusion). The major improvements come from the prompt design and this is an interesting finding as it shows that with careful prompt design we are able to outperform the more complex method that underwent extensive training.

We evaluate component contributions in our retrieval pipeline on 100 ICLR submissions (Table~\ref{tab:retrieval_ablation}).
The full pipeline combines keyword search with cited papers, ranks results by SPECTER2 embedding similarity, and applies GPT-3.5 reranking.
Without LLM reranking (embeddings only), we achieve 71\% overlap at top-10 with the full pipeline.
When considering only keyword search (excluding citations), overlap drops to 32\%, indicating that author citations provide crucial relevance signals beyond keyword matching.


\section{Conclusion}
We present a human-informed pipeline for automated novelty assessment in peer review, addressing a critical gap in AI-assisted review systems. Our approach combines systematic related work retrieval with structured evaluation criteria derived from analysis of expert reviewer patterns. Experimental results demonstrate that our system outperforms existing AI baselines and achieves higher agreement rates than human-human comparisons across key evaluation dimensions. 

\textcolor{black}{Our approach demonstrates that careful prompt design can achieve strong performance without requiring extensive model training. This is a strength of our method. While methods that attempt training \cite{zhu2025deepreviewimprovingllmbasedpaper, idahl-ahmadi-2025-openreviewer} require substantial computational resources (e.g., 8× H100 80G GPUs for 23,500 steps at 256K context), whereas our prompt-based approach achieves strong performance while offering greater computational efficiency.}


\section*{Limitations}

Despite achieving strong performance, our system has several important limitations:

\textbf{Evaluation Scope}: Our evaluation focuses on computer science papers from ICLR 2025. The system's performance on other scientific domains remains untested and likely requires domain-specific adaptations.

\textbf{Consistency vs. Diversity}: While our analysis shows that systematic evaluation reduces reviewer disagreement, this consistency might eliminate valuable diversity in perspectives. The 35-40\% human-human disagreement rate may reflect legitimate differences in expertise and viewpoint rather than mere inconsistency.

\textbf{Nuanced Novelty}: Breakthrough ideas often challenge conventional evaluation criteria. Our system's consistent approach might miss paradigm-shifting contributions that human experts would recognize through intuition or deep domain expertise.

\textbf{Language Scope}: Our study evaluates the system only on English‐language manuscripts and reviews. As a result, we cannot claim that the approach generalizes to submissions written in other languages or rooted in different academic conventions; assessing cross-lingual performance remains future work.

\textcolor{black}{\textbf{Human Analysis for Prompt Design}: We acknowledge our approach of analysis of the selected data lacks formal inter-rater reliability measures, but argue it was appropriate for this initial investigation into an understudied phenomenon, with the patterns’ effectiveness ultimately validated through our pipeline results.}

\textcolor{black}{\textbf{Human Study}: Our human evaluation is based on 100 pairwise comparisons with three expert annotators, comparable to related work in peer review analysis (\cite{Yuan_Liu_2022}: 40 papers; \cite{chamoun-etal-2024-automated} 100 examples; \cite{dycke-etal-2025-stricta}: 20+ papers). While a larger sample would provide additional confidence, novelty assessment requires in-depth domain expert annotators with deep familiarity with the relevant literature, making extensive human evaluation resource-prohibitive.}

\section*{Acknowledgments}
This work has been funded by the European Union (ERC, InterText, 101054961). Views and opinions expressed are however those of the author(s) only and do not necessarily reflect those of the European Union or the European Research Council. Neither the European Union nor the granting authority can be held responsible for them. This work has also been co-funded by the LOEWE Distinguished Chair “Ubiquitous Knowledge Processing”, LOEWE initiative, Hesse, Germany (Grant Number: LOEWE/4a//519/05/00.002(0002)/81).

\bibliography{custom}

\appendix
\label{sec:appendix}

\input{tables/depth}
\input{tables/priorwork}
\input{tables/retrieval_ablation}

\section{Data Analysis}

\subsection{Sampling Methodology}
\textcolor{black}{Our sampling is sentiment-agnostic. We sample for novelty discussions (both positive and negative), not specifically novelty issues. Our keywords ("novel", "contribution", "prior work", etc.) appear in both types of assessments.
\paragraph{Empirical analysis of sentiment distribution} We analyzed the sentiment of novelty discussions in our dataset (352 reviews total) and found: 45 Positive, 234 Negative, 73 Mixed. While negative discussions are more prevalent (as expected, since issues receive more attention in reviews), our data includes substantial coverage of positive and mixed novelty assessments, demonstrating that our sampling captures the full spectrum of novelty discussions rather than being biased toward issues only.}

\section{Human Evaluation Protocol: Novelty Assessment Comparison}
\label{sec:human-eval-protocol}

\subsection{Task Design}
We conducted a comparative evaluation where human evaluators assessed the quality of AI-generated novelty assessments against expert-written reference assessments. Each evaluator compared pairs of AI-generated assessments (labeled A and B) to a human expert's gold-standard novelty review of the same research paper.

\subsection{Evaluation Framework}

\paragraph{Materials Provided}
For each evaluation, evaluators received: (1) an expert-written gold-standard novelty review as reference, (2) two novelty assessments (A and B) with system identities hidden.

\paragraph{Evaluation Dimensions} Evaluators assessed each pair across four dimensions. For each dimension, evaluators selected one of four options: \textit{A wins}, \textit{B wins}, \textit{Tie} (both equally good/poor), or \textit{Unclear} (cannot determine).:

\begin{enumerate}
    \item \textbf{Reasoning Alignment:} Which assessment better captures the key novelty reasoning from the reference? Evaluators considered similarity of novelty claims, logical arguments, and focus areas.
    \item \textbf{Decision Alignment:} Which assessment reaches a novelty verdict most consistent with the reference? This included agreement on overall judgment (novel/incremental/mixed) and similar weighting of novelty factors.
    \item \textbf{Claim Substantiation:} Which assessment better supports its novelty claims with evidence? Evaluators looked for specific citations, concrete examples from the paper, and absence of unsupported generalizations.
    \item \textbf{Analytical Quality:} Which assessment provides more insightful technical analysis of novelty? This considered depth of technical discussion, specificity of analysis, and balanced consideration of strengths and limitations.
\end{enumerate}

\subsection{Evaluation Guidelines}

\paragraph{Instructions for Evaluators}
Evaluators were instructed to read the reference assessment thoroughly before evaluating A and B, evaluate each dimension independently, and base judgments on substantive content rather than stylistic differences. They allocated 4--7 minutes per example to ensure thorough evaluation and flagged ambiguous cases with explanatory comments when necessary.

\paragraph{Evaluation Focus}
Evaluators were instructed to \textbf{prioritize} substance and accuracy of novelty reasoning, alignment with reference judgments (particularly for Dimensions 1--2), quality and depth of technical analysis (particularly for Dimensions 3--4), and specific evidence and citations supporting claims. They were instructed to \textbf{disregard} writing style, grammar, or formatting differences; suggestions for paper improvement unrelated to novelty; minor phrasing variations with equivalent meaning; and length differences if content quality was comparable.

\subsection{Implementation Details}

\begin{figure*}[h]
    \centering
    \includegraphics[width=\textwidth]{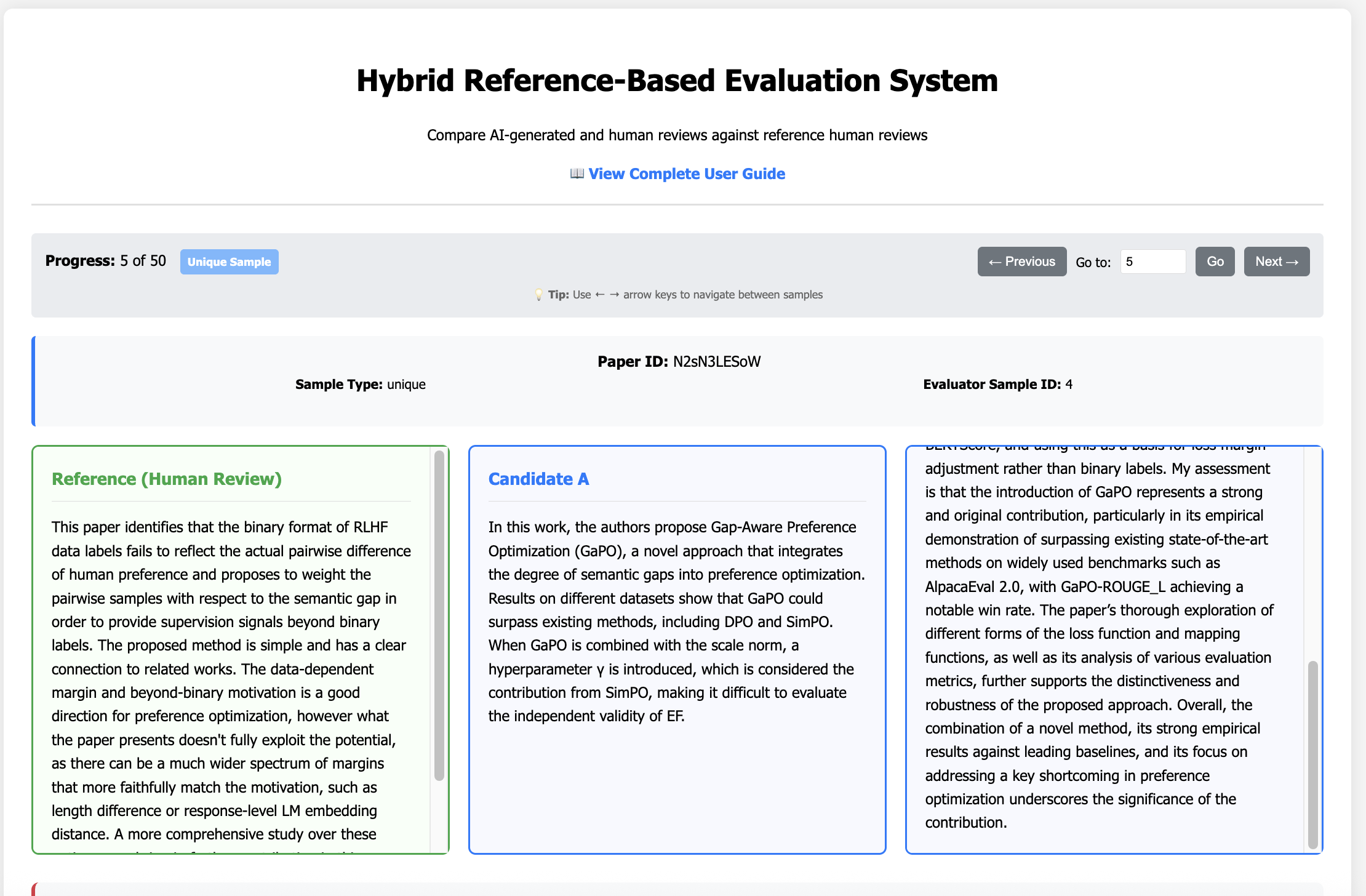}
    \caption{Screenshot of the custom-built interface used for human evaluation. Annotators compared AI-generated and human-written novelty assessments across multiple dimensions, including reasoning depth, prior work engagement, and conclusion alignment.}
    \label{fig:human-eval-ui}
\end{figure*}

\begin{figure*}[h]
    \centering
    \includegraphics[width=\textwidth]{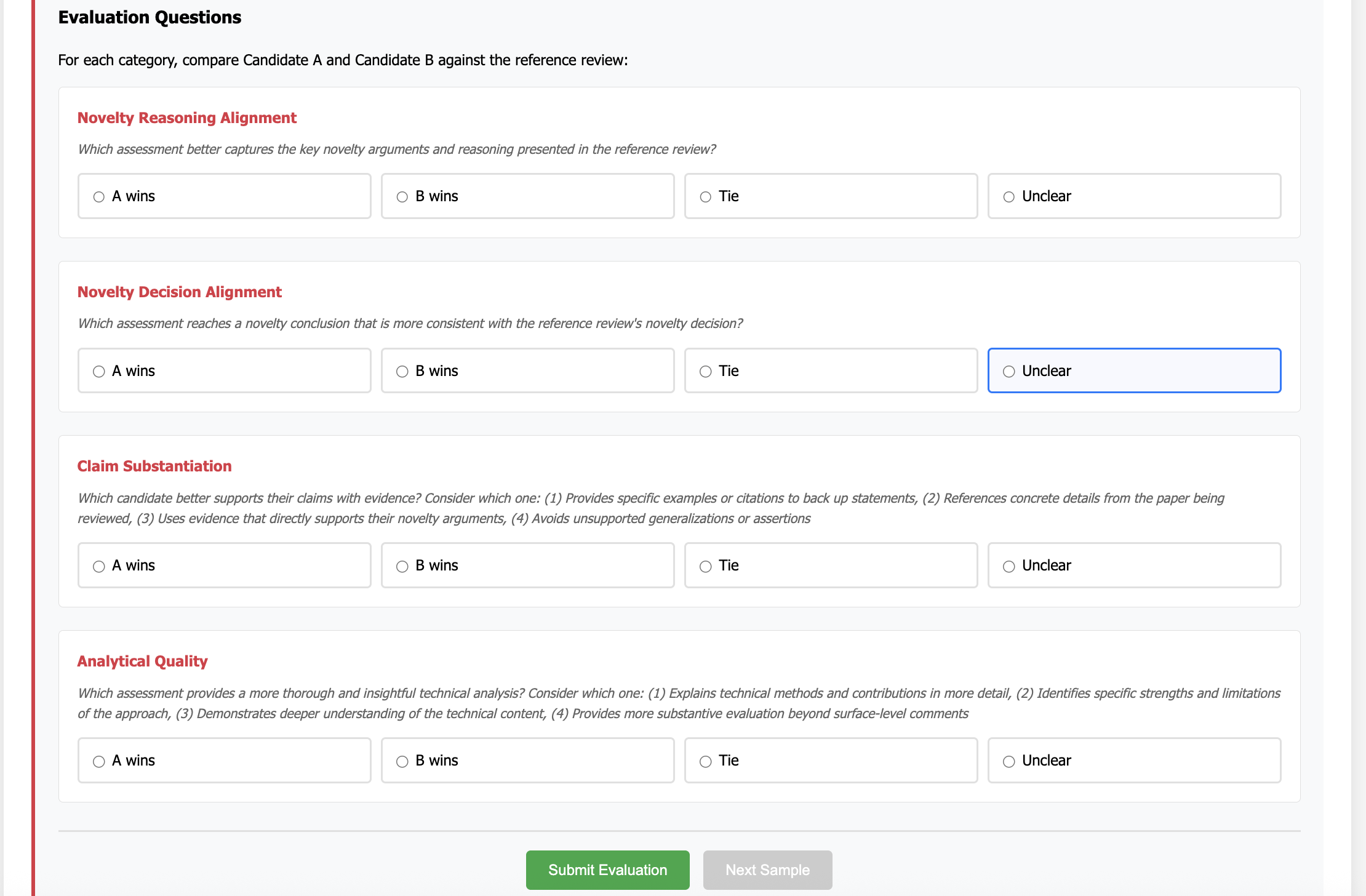}
    \caption{Screenshot (2) of the custom-built interface used for human evaluation. Annotators compared AI-generated and human-written novelty assessments across multiple dimensions, including reasoning depth, prior work engagement, and conclusion alignment.}
    \label{fig:human-eval-ui-2}
\end{figure*}

\paragraph{Evaluation Platform}
The evaluation was conducted through a custom web interface presenting materials in a standardized format (see Figures~\ref{fig:human-eval-ui} and \ref{fig:human-eval-ui-2}). Each evaluator received a unique evaluator ID, 50 randomly assigned paper-assessment pairs, and the ability to save progress and flag unclear cases.

\paragraph{Quality Control}
We calculated inter-evaluator agreement using Cohen's kappa reported in Table \ref{tab:irr_metrics}.

\paragraph{Data Collection}
Completed evaluations were submitted as structured JSON files containing dimension-wise selections (A/B/Tie/Unclear), time spent per evaluation, and comments for flagged cases.

\input{tables/IAA}

\section{Output Examples}

Output of our pipeline can be seen in Tables \ref{tab:full-text-comparison}, \ref{tab:durnd-comparison} and \ref{tab:meta-instructions-comparison}. It is quite evident that our system aligns better with the human as compared to the baselines across all four dimensions.

\section{Understanding Human alignment patterns}
\label{sec:analysis}

\textcolor{black}{\paragraph{Pattern Analysis} We analyzed 45 papers where multiple human reviewers reached different novelty conclusions. This was determined from the LLM as judge results we received during evaluation. In a human-AI collaborative setup, we first iteratively examined review pairs to identify recurring disagreement patterns, then developed categories along two observable dimensions: (1) focus divergence - what aspects reviewers discussed (methods, applications, results, prior work, etc.), and (2) assessment granularity - their level of analytical detail (high-level vs detailed). We then used Claude Code to perform side-by-side comparative reading and categorization of all 45 review pairs according to these predefined categories. Our analysis revealed the patterns: 62.2\% of cases (28/45) showed granularity differences with one reviewer providing detailed component-level analysis while another gave high-level assessment, and 75.6\% of cases (34/45) showed focus differences with reviewers evaluating different aspects of the work.}
\textcolor{black}{\paragraph{Misrepresentation of Novelty} Additionally we manually reviewed 10 generated outputs from the novelty-delta-analysis stage to  interpret where does the misrepresentation of novelty arise from, we analyzed the structure and reasoning patterns. We found that the system's primary mode of analysis involves evaluating how authors characterize their contributions relative to cited works, using the citation contexts from the paper. When relevant uncited work is identified, the system flags it for additional comparison but bases its core novelty assessment on the cited literature where authors' explicit positioning is available.
This pattern reveals that most identifiable novelty overstatement arises from inadequate differentiation from already-cited work. Because we have access to authors' own citation contexts, we can directly evaluate whether their novelty claims hold up against how they characterized prior work. In contrast, for uncited works, we can identify potential gaps but lack the authors' explicit framing of the novelty relationship, making these better suited for clarification during rebuttal rather than definitive assessment.
}

\textcolor{black}{
\paragraph{Factuality Analysis} LLMs are known to hallucinate references, which is precisely why our pipeline is specifically designed to be grounded through an extensive multi-step retrieval pipeline with multiple reranking stages. Each related paper alongside the submission is also used in our pipeline.
To directly address whether our system’s citations accurately support the conclusions or occasionally introduce factual inconsistencies, we conducted a systematic manual verification study.}

\textcolor{black}{We randomly sampled 50 factual claims from our novelty delta analysis outputs. Each claim was manually verified against the actual paper. Claims were categorized as ACCURATE, PARTIALLY-ACCURATE, INACCURATE, or CANNOT-VERIFY. 
As can be seen in table \ref{tab:verification-status} that 96\% of claims were accurate or partially accurate. The 14 partially accurate claims typically involved minor discrepancies such as incorrect author attribution (e.g., "Sun et al." instead of "Xu et al.") or year errors (e.g., 2024 instead of 2023), while the core method descriptions remained correct. Only 2 claims (4\%) contained substantive factual errors. One misattributing a paper's domain (text vs. image attacks) and one overstating a method's historical significance.t}
\begin{table}[h]
\centering
\begin{tabular}{lcc}
\hline
\textbf{Verification Status} & \textbf{Count} & \textbf{Percentage} \\
\hline
Accurate            & 34 & 68\% \\
Partially Accurate  & 14 & 28\% \\
Inaccurate          & 2  & 4\%  \\
\hline
\end{tabular}
\caption{Verification Status Summary}
\label{tab:verification-status}
\end{table}

\definecolor{verdict-negative}{RGB}{255, 230, 230}
\definecolor{verdict-positive}{RGB}{220, 255, 220}
\definecolor{verdict-neutral}{RGB}{220, 220, 255}
\definecolor{critical}{RGB}{255, 200, 200}

\newcommand{\wrapcolorbox}[2]{%
  \colorbox{#1}{\parbox[t]{\dimexpr\linewidth-2\fboxsep}{#2}}%
}

\begin{table*}[t]
\centering
\small
\begin{tabular}{>{\raggedright\arraybackslash}p{0.28\textwidth} 
                | >{\raggedright\arraybackslash}p{0.28\textwidth} 
                | >{\raggedright\arraybackslash}p{0.36\textwidth}}
\toprule
\textbf{Human (Reference)} & \textbf{Scideator (Baseline)} & \textbf{Ours (Proposed)} \\
\midrule
The proposed approach aims to significantly reduce memory consumption during long-sequence inference while maintaining model performance and requiring minimal modifications to existing LLM frameworks. While the integration of CGE and RGL is presented as a novel solution to the KV cache memory challenge, \wrapcolorbox{verdict-negative}{the overall novelty of the work is limited}. Sparse attention mechanisms have already been \wrapcolorbox{verdict-neutral}{extensively explored in prior literature, such as [1] and [2]}, which diminishes the originality of the proposed methods. Furthermore, the CGE component \wrapcolorbox{verdict-neutral}{closely resembles previous approaches like H20 [3] and SnapKV [4], with only incremental differences}. The feedback mechanism using sliding windows has also been well-studied in H20 [3]. As such, \wrapcolorbox{verdict-negative}{the main contribution appears to be a specific combination and implementation of existing ideas rather than a fundamentally new technique}. The paper would benefit from a clearer articulation of how its methods differ from these established approaches to better establish its unique contribution. 
& 
\wrapcolorbox{verdict-positive}{\texttt{- Class: novel}} \newline The idea is novel because IntelLLM introduces \wrapcolorbox{verdict-positive}{a distinct approach to key-value (KV) cache compression} by using strategies like center of gravity eviction (CGE) and remote gap localization (RGL), which are \wrapcolorbox{verdict-positive}{not seen in existing works} such as RazorAttention[o] and LeanKV[1]. These methods prioritize retaining essential tokens and leveraging positional features to enhance compression efficiency, offering \wrapcolorbox{verdict-positive}{a dedicated approach with unique techniques} for balancing compression rate and performance.
& 
IntelLLM introduces new heuristics—Center of Gravity Eviction (CGE) and Remote Gap Localization (RGL)—for token selection and long-range dependency preservation in KV cache compression for LLM inference, but \wrapcolorbox{verdict-negative}{these are incremental variants of established token eviction approaches}. The submission \wrapcolorbox{critical}{overstates its novelty}, as \wrapcolorbox{verdict-neutral}{the core ideas (token selection, attention sparsity, training-free deployment) are already well-explored}, and similar methods (e.g., RazorAttention, PyramidKV, L$_2$ Norm) achieve comparable goals without model changes or fine-tuning. \wrapcolorbox{critical}{Several highly relevant recent works are omitted from the discussion}, and \wrapcolorbox{critical}{the claims of being the first to balance compression and performance or to preserve long-range dependencies are not substantiated by the literature}. \wrapcolorbox{verdict-negative}{The main technical delta lies in the specific heuristics (CGE, RGL) and their empirical performance, rather than in a conceptual advance}. Reviewers should view IntelLLM as \wrapcolorbox{verdict-negative}{a routine, incremental contribution} and may wish to request more comprehensive comparisons and a more accurate positioning within the current research landscape. \\
\bottomrule
\end{tabular}
\caption{Full novelty assessments from the human reviewer (reference), the Scideator baseline, and our proposed system. Key phrases are highlighted to show verdict alignment: \colorbox{verdict-positive}{positive novelty claims}, \colorbox{verdict-negative}{limited/incremental novelty}, \colorbox{verdict-neutral}{comparative analysis}, and \colorbox{critical}{critical issues}.}
\label{tab:full-text-comparison}
\end{table*}

\definecolor{verdict-negative}{RGB}{255, 230, 230}
\definecolor{verdict-positive}{RGB}{220, 255, 220}
\definecolor{verdict-neutral}{RGB}{220, 220, 255}
\definecolor{critical}{RGB}{255, 200, 200}

\begin{table*}[t]
\centering
\small
\begin{tabular}{>{\raggedright\arraybackslash}p{0.28\textwidth} 
                | >{\raggedright\arraybackslash}p{0.28\textwidth} 
                | >{\raggedright\arraybackslash}p{0.36\textwidth}}
\toprule
\textbf{Human (Reference)} & \textbf{DeepReviewer (Baseline)} & \textbf{Ours (Proposed)} \\
\midrule
The paper "DuRND: Rewarding from Novelty to Contribution for Reinforcement Learning via Dual Random Networks Distillation" proposes \wrapcolorbox{verdict-neutral}{an extension to classical RND} by introducing two distinct random network modules—one for states deemed "successful" and another for states associated with "failure." This innovation allows for the derivation of both a "novelty" and a "contribution" reward signal, \wrapcolorbox{verdict-positive}{striking a balance between exploratory and exploitative behavior}. While \wrapcolorbox{verdict-negative}{the additional novelty introduced by DuRND is incremental compared to classical RND}, I still believe \wrapcolorbox{verdict-positive}{the contribution is valuable and fills a gap in the current literature}.
& 
This paper proposes \wrapcolorbox{verdict-positive}{a novel framework, Dual Random Networks Distillation (DuRND)}, to address the exploration-exploitation dilemma in reinforcement learning by combining novelty-based exploration and value-based exploitation in a unified framework. The novelty and contribution rewards in DuRND are based on prediction errors from random network modules, which may not capture complex state relationships in certain environments, and the reliance on prediction error as a proxy for novelty might be insufficient where the state space has a complex structure or the prediction error does not align with actual information gain. The approach introduces \wrapcolorbox{verdict-positive}{an interesting combination of novelty and contribution rewards} but \wrapcolorbox{verdict-neutral}{relies on computationally efficient heuristics that might not capture the true novelty in complex environments}. While the framework demonstrates effectiveness and low computational overhead, \wrapcolorbox{verdict-negative}{the linear adjustment of weights for novelty and contribution rewards is a heuristic that may not be optimal or universally applicable}. Overall, \wrapcolorbox{verdict-positive}{the originality lies in the integration of novelty and value-based signals}, but there remain \wrapcolorbox{verdict-negative}{significant limitations in the robustness and generalizability of the contribution} due to dependency on specific proxy measures and heuristic schedules.
& 
DuRND is \wrapcolorbox{verdict-neutral}{a direct extension of Random Network Distillation (RND)}, introducing a dual random network architecture to compute both novelty (exploration) and "contribution" (exploitation) rewards, with a dynamic scheduling mechanism that shifts emphasis over time. While the use of random networks for both signals is \wrapcolorbox{verdict-positive}{a novel implementation detail}, \wrapcolorbox{verdict-negative}{the underlying concepts—combining exploration and exploitation signals and dynamically adjusting their balance—are well-established in the literature} (e.g., SORS, ROSA, Sibling Rivalry, SERENE). The authors accurately position DuRND as a low-overhead, scalable method, but \wrapcolorbox{critical}{somewhat overstate its conceptual novelty and the dichotomy between prior approaches}. Empirical results show improved performance on standard sparse-reward benchmarks, though \wrapcolorbox{verdict-neutral}{such gains may be attributable to implementation choices rather than fundamental innovation}. Overall, DuRND offers \wrapcolorbox{verdict-negative}{a practical and well-engineered architectural tweak to RND}, with its main contribution being \wrapcolorbox{verdict-negative}{a simple, dual-network design for joint intrinsic reward computation, rather than a conceptual breakthrough}. \\
\bottomrule
\end{tabular}
\caption{Full novelty assessments from the human reviewer (reference), the DeepReviewer baseline, and our proposed system for the DuRND paper. Key phrases are highlighted to show verdict alignment: \colorbox{verdict-positive}{novel/valuable aspects}, \colorbox{verdict-negative}{limited/incremental novelty}, \colorbox{verdict-neutral}{technical details}, and \colorbox{critical}{overstated claims}.}
\label{tab:durnd-comparison}
\end{table*}

\definecolor{verdict-negative}{RGB}{255, 230, 230}
\definecolor{verdict-positive}{RGB}{220, 255, 220}
\definecolor{verdict-neutral}{RGB}{220, 220, 255}
\definecolor{critical}{RGB}{255, 200, 200}

\begin{table*}[t]
\centering
\small
\begin{tabular}{>{\raggedright\arraybackslash}p{0.28\textwidth} 
                | >{\raggedright\arraybackslash}p{0.28\textwidth} 
                | >{\raggedright\arraybackslash}p{0.36\textwidth}}
\toprule
\textbf{Human (Reference)} & \textbf{OpenReviewer (Baseline)} & \textbf{Ours (Proposed)} \\
\midrule
This paper introduces a method for generating image inputs to Vision Language Models (VLMs) that embed "meta-instructions," enabling the model to respond to any user query appended to the image with a specified "spin," such as a particular sentiment or language. While the concept of providing meta-instructions through image inputs is \wrapcolorbox{verdict-negative}{not entirely novel—prior work, notably Bailey et al. (2023), has explored related prompt injection attacks via images} this paper offers the most thorough and comprehensive treatment of the subject to date. The general algorithm employed, which involves creating a dataset of input-output pairs and training an image via projected gradient descent to maximize the likelihood over this dataset, \wrapcolorbox{verdict-neutral}{closely aligns with the "Behavior Matching" approach described by Bailey et al. (2023)}. As such, \wrapcolorbox{verdict-negative}{the methodological contribution is incremental rather than fundamentally new}. However, the paper distinguishes itself through \wrapcolorbox{verdict-positive}{its in-depth analysis and experimental results}, particularly the study of semantic changes in images resulting from various attacks, with a focus on how meta-instruction attacks can preserve the original image meaning. The transferability experiments and the breadth of evaluation across multiple VLMs further strengthen the contribution. Overall, \wrapcolorbox{verdict-negative}{the novelty of this work lies less in the method itself and more in the scope, depth, and clarity of its empirical investigation}, as well as the new insights it provides into the vulnerabilities of VLMs to image-based prompt injection. The paper would benefit from more explicitly acknowledging the overlap with prior methods and more clearly articulating how its results advance the understanding of this threat model.
& 
This paper introduces \wrapcolorbox{verdict-positive}{a novel method of embedding hidden meta-instructions in images} as a way to attack visual language models, which is \wrapcolorbox{verdict-positive}{distinct from traditional jailbreaking or adversarial example attacks}. While the approach is new, the paper could provide more insight into how the proposed attacks compare to existing forms of adversarial attacks on VLMs, highlighting the unique aspects of meta-instruction attacks.
& 
This assessment finds that the submission addresses indirect, cross-modal prompt injection in Visual Language Models (VLMs) by embedding hidden meta-instructions in images, aiming to steer model outputs while preserving image semantics. The work is \wrapcolorbox{verdict-neutral}{most closely related to recent studies on adversarial image prompting (e.g., Qi et al. 2024, Bagdasaryan et al. 2023)}, but distinguishes itself through \wrapcolorbox{verdict-positive}{more systematic optimization for semantic preservation and a broader range of meta-instructions beyond jailbreaking}. The main substantive contributions are \wrapcolorbox{verdict-positive}{a rigorous, multi-metric evaluation of attack effectiveness and semantic preservation}, and empirical evidence that image-based meta-instructions can be more effective than explicit text prompts. However, the assessment notes that \wrapcolorbox{verdict-negative}{the conceptual advances are incremental}, as \wrapcolorbox{verdict-neutral}{the core idea of cross-modal prompt injection and semantic preservation has been explored in prior work}, and \wrapcolorbox{critical}{some novelty claims (e.g., being the first to frame VLM users as victims) are somewhat overstated}. Overall, \wrapcolorbox{verdict-negative}{the submission's primary strengths lie in evaluation rigor and empirical findings, while its conceptual contributions represent a natural progression of the field rather than a fundamental shift}. \\
\bottomrule
\end{tabular}
\caption{Full novelty assessments from the human reviewer (reference), the OpenReviewer baseline, and our proposed system for the Meta-Instructions in VLMs paper. Key phrases are highlighted to show verdict alignment: \colorbox{verdict-positive}{novel/strength claims}, \colorbox{verdict-negative}{limited/incremental novelty}, \colorbox{verdict-neutral}{prior work comparison}, and \colorbox{critical}{overstated claims}.}
\label{tab:meta-instructions-comparison}
\end{table*}

\begin{figure}[h]
\centering
\includegraphics[width=0.95\columnwidth]{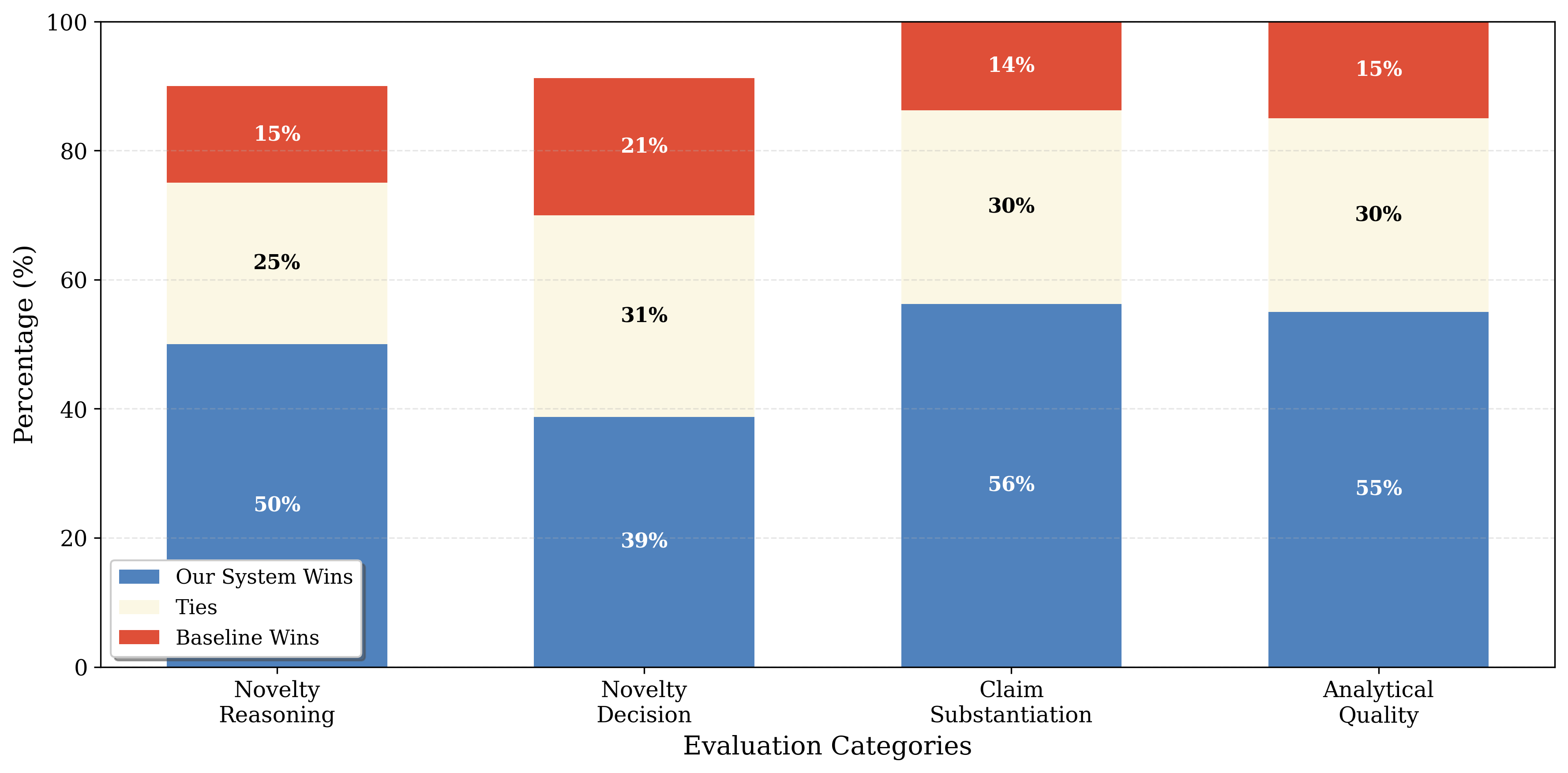}
\caption{ Performance breakdown across evaluation categories, aggregated across all baseline comparisons.}
\label{fig:human_eval_dim}
\end{figure}

\begin{figure}[h]
    \centering
    \includegraphics[width=0.9\columnwidth]{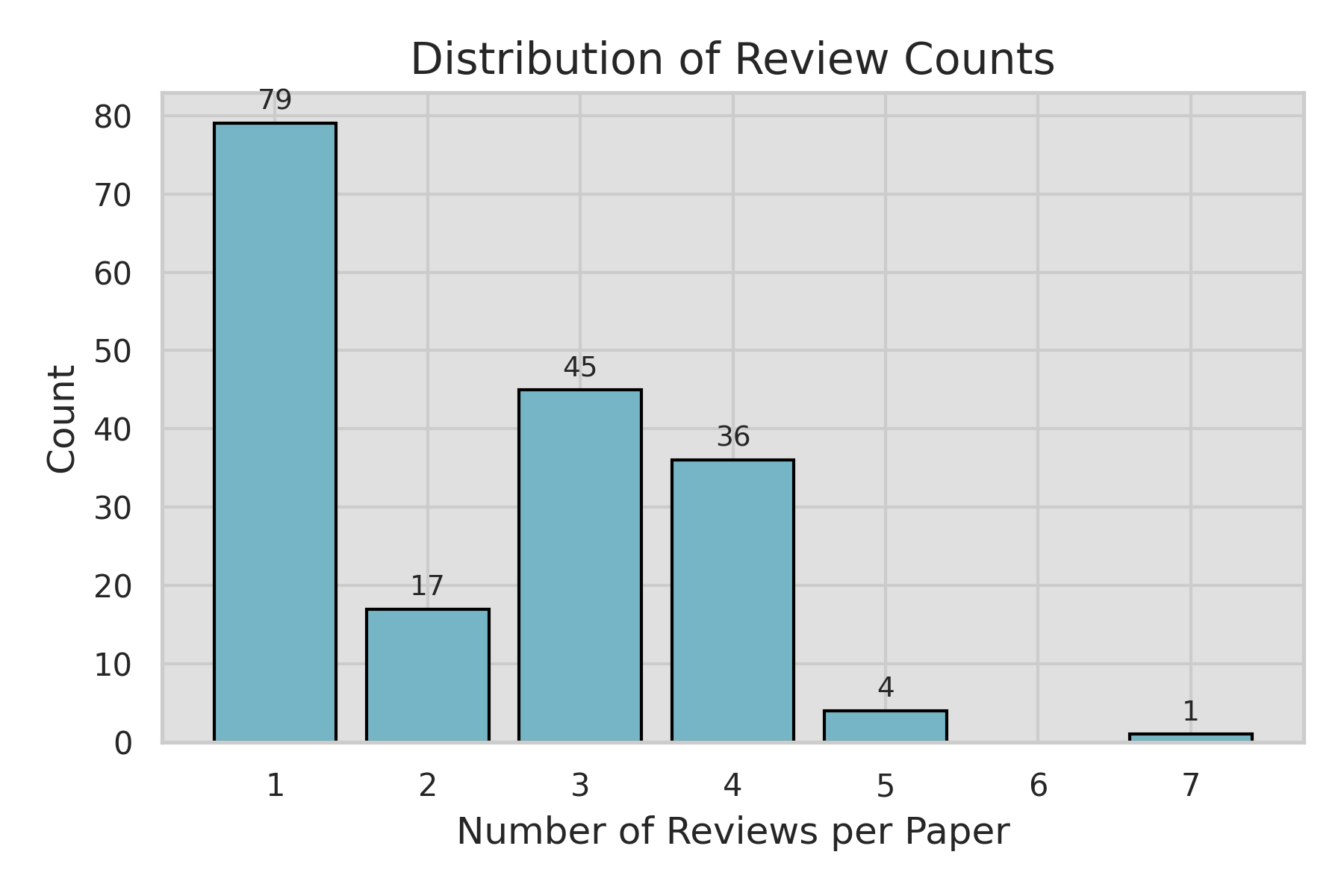}
    \caption{Distribution of the number of reviews per paper. Most papers received 1 to 4 reviews.}
    \label{fig:review_count_dist}
\end{figure}

\input{Figures/prompt_str_out}

\end{document}

%% file: tables/dataset_count.tex
\begin{table}[t]
\centering
\scriptsize
\setlength{\tabcolsep}{3pt}
\begin{tabular}{lrrrr}
\toprule
Decision & Papers & Reviews & Words/rev & Rev/paper \\
\midrule
No Decision / Withdrawn & 51 & 110 & 1002 & 2.16 \\
Reject                  & 81 & 195 & 919  & 2.41 \\
Accept (Poster)         & 45 & 102 & 962  & 2.27 \\
Accept (Spotlight)      & 4  & 10  & 997  & 2.50 \\
Accept (Oral)           & 1  & 2   & 1182 & 2.00 \\
\midrule
Total                   & 182 & 419 & 959 & 2.30 \\
\bottomrule
\end{tabular}
\caption{Distribution of papers and reviews with novelty discussions by ICLR 2025 decision outcomes}
\label{tab:dataset_stats}
\end{table}

%% file: tables/JudgementEval.tex
\begin{table*}[t]
  \centering
  \resizebox{\textwidth}{!}{%
    \begin{tabular}{lcccc}
      \toprule
      \textbf{System} &
      \textbf{Reasoning Alignment (\%↑)} &
      \textbf{Conclusion Agreement (\%↑)} &
      \textbf{Positive Shift (\%↓)} &
      \textbf{Negative Shift (\%↓)} \\
      \midrule
      OpenReviewer \cite{idahl-ahmadi-2025-openreviewer}
                      & 42.4 $\pm$ 0.39 & 46.8 $\pm$ 0.71 & 6.3 $\pm$ 0.27 & 15.3 $\pm$ 0.40 \\
      DeepReviewer \cite{zhu2025deepreviewimprovingllmbasedpaper}
                      & 50.6 $\pm$ 0.67 & 51.5 $\pm$ 1.24 & 21.7 $\pm$ 1.89 & 9.1 $\pm$ 0.00 \\
      Human vs.\ Human
                      & 65.1 $\pm$ 1.05 & 62.8 $\pm$ 0.40 & 6.7 $\pm$ 0.79 & 15.0 $\pm$ 0.40 \\
      Scideator \cite{radensky2025scideatorhumanllmscientificidea}
                      & 23.7 $\pm$ 0.00 & 22.4 $\pm$ 0.00 & 0.0 $\pm$ 0.00 & 20.5 $\pm$ 0.00 \\
      \textbf{Ours (GPT-4.1)}
                      & \textbf{86.5 $\pm$ 0.20} & \textbf{75.3 $\pm$ 0.85} & 16.3 $\pm$ 1.28 & 3.0 $\pm$ 0.43 \\
      \bottomrule
    \end{tabular}%
  }
  \caption{Summary of Reasoning Alignment, Conclusion Agreement, Positive Shift, and Negative Shift Metrics}
  \label{tab:overall_summary}
\end{table*}

%% file: tables/ablations.tex
\begin{table}[t]
  \centering
  \resizebox{\columnwidth}{!}{%
    \begin{tabular}{lcc}
      \toprule
      \textbf{System Configuration} &
      \textbf{Reasoning (\%)} &
      \textbf{Conclusion (\%)} \\
      \midrule
      Naive Prompt & 39.3 & 24.7 \\
      + Our Prompt Design & 80.0 (+40.7) & 71.5 (+46.8) \\
      + Structured Extraction & 83.3 (+3.3) & 76.0 (+4.5) \\
      + Landscape Analysis (Full) & 86.5 (+3.2) & 75.3 (–0.7) \\
      \bottomrule
    \end{tabular}%
  }
    \caption{Component Analysis: Incremental Contribution of Pipeline Components}
    \label{tab:component_analysis}
\end{table}

%% file: tables/depth.tex
\begin{table}[t]
  \centering
  \resizebox{\columnwidth}{!}{%
    \begin{tabular}{lccc}
      \toprule
      \textbf{System} & \textbf{Surface-Level (\%)} & \textbf{Moderate (\%)} & \textbf{Deep (\%)} \\
      \midrule
      OpenReviewer   & 67.4 & 31.3 & 1.2 \\
      DeepReviewer   & 43.4 & 56.6 & 0.0 \\
      Human vs.\ Human & 22.3 & 66.2 & 11.5 \\
      Scideator & 44.9 & 54.5 & 0.6 \\
      \textbf{Ours}  & 0.0 & 47.9 & 52.1 \\
      \bottomrule
    \end{tabular}%
  }
    \caption{Reasoning Depth Distribution (Percentages)}
      \label{tab:depth}
  \end{table}


%% file: tables/priorwork.tex

\begin{table}
  \resizebox{\columnwidth}{!}{%
    \begin{tabular}{lccc}
      \toprule
      \textbf{System} & \textbf{None (\%)} & \textbf{Limited (\%)} & \textbf{Extensive (\%)} \\
      \midrule
      OpenReviewer   & 39.9 & 53.1 & 7.0 \\
      DeepReviewer   & 24.7 & 75.3 & 0.0 \\
      Human vs.\ Human & 19.6 & 65.2 & 15.2 \\
      Scideator & 0.0 & 75.9 & 24.1 \\
      \textbf{Ours}  & 0.0 & 39.1 & 60.9 \\
      \bottomrule
    \end{tabular}%
  }
    \caption{Prior Work Engagement Distribution (Percentages)}
    \label{tab:priorwork}
\end{table}

%% file: tables/retrieval_ablation.tex
\begin{table}[t]
  \centering
  \resizebox{\columnwidth}{!}{%
    \begin{tabular}{lcccc}
      \toprule
      \textbf{Method} &
      \textbf{Top-5} &
      \textbf{Top-10} &
      \textbf{Top-15} &
      \textbf{Top-20} \\
      \midrule
      Full Pipeline & 1.000 & 1.000 & 1.000 & 1.000 \\
      Dense Only & 0.612 $\pm$ 0.20 & 0.710 $\pm$ 0.18 & 0.748 $\pm$ 0.12 & 0.766 $\pm$ 0.10 \\
      KW Only & 0.300 $\pm$ 0.27 & 0.317 $\pm$ 0.24 & 0.345 $\pm$ 0.22  & 0.349 $\pm$ 0.19 \\
      \bottomrule
    \end{tabular}%
  }
  \caption{Retrieval Pipeline Ablation Study}
  \label{tab:retrieval_ablation}
\end{table}

%% file: tables/IAA.tex
\begin{table}[h]
  \centering
  \resizebox{\columnwidth}{!}{%
    \begin{tabular}{lccc}
      \toprule
      \textbf{Category} & \textbf{Agreement} & \textbf{Kappa} & \textbf{Comparisons} \\
      \midrule
      Novelty Reasoning Alignment & 0.520 & 0.341 & 75 \\
      Novelty Decision Alignment  & 0.533 & 0.346 & 75 \\
      Claim Substantiation        & 0.493 & 0.287 & 75 \\
      Analytical Quality          & 0.560 & 0.368 & 75 \\
      \bottomrule
    \end{tabular}%
  }
    \caption{Inter-Rater Reliability Metrics Across Categories}
    \label{tab:irr_metrics}
\end{table}

%% file: Figures/prompt_str_out.tex
\definecolor{codebg}{gray}{0.95}

\begin{figure*}[t]
\centering
\begin{tcolorbox}[
    enhanced,
    width=\textwidth,
    colback=white,
    colframe=black,
    boxrule=0.8pt,
    arc=4pt,
    title=\textbf{Research Paper Information Extraction Prompt},
    coltitle=white,
    colbacktitle=black,
    toptitle=1mm,
    bottomtitle=1mm,
    fonttitle=\bfseries,
]
\lstset{
    basicstyle=\ttfamily\small,
    breaklines=true,
    frame=none,
    columns=fullflexible,
    keepspaces=true
}
\begin{lstlisting}
You are tasked with extracting key information from a research paper 
for building a knowledge representation.
Paper title: {title}
Based on the paper content provided below, extract the following information:
- "methods": [List of methods/approaches proposed in the paper],
- "problems": [List of problems the paper addresses],
- "datasets": [List of datasets used for evaluation],
- "metrics": [List of evaluation metrics used],
- "results": [List of objects with 'metric' and 'value' fields 
  representing key quantitative results],
- "novelty_claims": [Claims about what is novel in this work]
Be precise and specific.
Paper content:
{abstract}
{introduction}
\end{lstlisting}
\end{tcolorbox}
\caption{Research Paper Information Extraction Prompt}
\label{fig:extraction-prompt}
\end{figure*}

\begin{figure*}[t]
\centering
\begin{tcolorbox}[
  enhanced,
  width=\textwidth,          
  colback=white, 
  colframe=black, 
  boxrule=0.8pt, 
  arc=4pt,
  title=\textbf{Research Landscape Analysis},
  coltitle=white, 
  colbacktitle=black, 
  toptitle=1mm, 
  bottomtitle=1mm, 
  fonttitle=\bfseries,
]
\lstset{
    basicstyle=\ttfamily\scriptsize, 
    breaklines=true,
    frame=none,
    columns=fullflexible,
    keepspaces=true
}
\begin{lstlisting}
# Research Landscape Analysis

## Task
Analyze the collection of research papers provided below to create a 
comprehensive map of the research landscape they represent. The submission 
paper is the focus of our analysis, and the related papers provide context.

## Input Format
You will be provided with structured information extracted from multiple 
research papers including:
- A submission paper that is the focus of our analysis
- Multiple related papers that form the research context

Each paper contains:
- methods: List of methods/approaches proposed
- problems: List of problems addressed
- datasets: List of datasets used
- metrics: List of evaluation metrics
- results: Key quantitative results
- novelty_claims: Claims about what is novel in the work

## Output Format
Provide a comprehensive analysis with the following sections:

1. METHODOLOGICAL LANDSCAPE
   - Identify and describe the main methodological approaches across the papers
   - Group similar or related methods into clusters
   - Highlight methodological trends or patterns
   - Describe relationships between different methodological approaches

2. PROBLEM SPACE MAPPING
   - Identify the key problems being addressed across the papers
   - Analyze how different papers approach similar problems
   - Highlight patterns in problem formulation

3. EVALUATION LANDSCAPE
   - Analyze the common datasets and evaluation methods
   - Identify patterns in how performance is measured
   - Compare evaluation approaches across papers

4. RESEARCH CLUSTERS
   - Identify groups of papers that appear closely related
   - Describe the key characteristics of each cluster
   - Analyze relationships between clusters

5. TECHNICAL EVOLUTION
   - Identify any visible progression or evolution of ideas
   - Highlight building blocks and their extensions
   - Note any competing or complementary approaches

## Example Output Format
METHODOLOGICAL LANDSCAPE
- Cluster 1: [Description of similar methods across papers]
  - Papers X, Y, Z employ transformer-based approaches with variations in...
  - These methods share characteristics such as...
  - They differ primarily in...

PROBLEM SPACE MAPPING
- Problem Area 1: [Description of a common problem addressed]
  - Papers A, B, C all address this problem but differ in...
  - The problem is formulated differently in Paper D which focuses on...

... [additional sections] ...

Ensure your analysis is comprehensive, identifying significant patterns 
and relationships across the collection of papers.

## Papers:
{papers}
\end{lstlisting}
\end{tcolorbox}
\caption{Research Landscape Analysis Prompt}
\label{fig:landscape-prompt}
\end{figure*}

\clearpage
\begin{figure*}[t]
\centering
\begin{tcolorbox}[
    enhanced,
    width=\textwidth,
    colback=white,
    colframe=black,
    boxrule=0.8pt,
    arc=4pt,
    title=\textbf{Novelty Delta Analysis for Reviewer Support - Part 1},
    coltitle=white,
    colbacktitle=black,
    toptitle=1mm,
    bottomtitle=1mm,
    fonttitle=\bfseries,
    breakable
]
\lstset{
    basicstyle=\ttfamily\small,
    breaklines=true,
    frame=none,
    columns=fullflexible,
    keepspaces=true
}
\begin{lstlisting}
# Novelty Delta Analysis for Reviewer Support

## Task
Independently analyze how the submission paper's contributions relate to existing 
work in the field, critically examining both author claims and actual relationships. 
This analysis should help reviewers assess novelty by providing objective comparisons 
with prior work.

## Input Format
You will be provided with:
1. The structured information from the submission paper
2. A comprehensive research landscape analysis
3. Citation sentences for key related papers (how authors cite and characterize these works)

## Key Analysis Principles
- Independently verify relationships between submission and prior work
- Critically examine how authors characterize and compare with prior work
- Identify discrepancies between author characterizations and actual relationships
- Present evidence-based observations without making final judgments
- Distinguish between author-claimed differences and independently observed differences
- Provide context about field maturity and related work

## Output Format
Provide a detailed analysis with the following sections:

1. RESEARCH CONTEXT POSITIONING
   - Situate the submission within the identified research landscape
   - Identify the most closely related prior works
   - Independently assess how the submission relates to existing methodological clusters
   - Analyze its place within the problem space and evaluation approaches
   - Note: Do not accept author positioning claims without verification

2. AUTHOR CITATION ANALYSIS
   - Analyze how authors characterize and compare with each cited related work
   - Identify patterns in how authors position their contributions relative to others
   - Assess whether characterizations of prior work are accurate and balanced
   - Note discrepancies between how authors describe prior work and independent assessment
   - Evaluate whether claimed improvements or differences are substantiated
   - Identify rhetoric that may overstate differences or understate similarities

3. CONTRIBUTION DELTA ANALYSIS
   For each main contribution claimed in the submission:
   - Identify the most similar prior work for this specific contribution
   - Critically examine whether claimed differences actually exist
   - Detail exactly how this contribution differs from prior work, based on evidence
   - Compare author characterizations with independently verified relationships
   - Distinguish between substantive differences and superficial variations
   - Note when author claims about novelty or extension may be overstated
   - Consider whether improvements might be due to implementation details rather than 
     conceptual advances
   - Note: Present factual observations about deltas without accepting author framing

4. FIELD CONTEXT CONSIDERATIONS
   - Provide information about how active/mature this research area is
   - Identify recent survey papers or literature reviews in this space
   - Note trends in how the field has been evolving
   - Present context about typical incremental advances in this field
   - Note: Offer context that helps reviewers calibrate their expectations
\end{lstlisting}
\end{tcolorbox}
\caption{Novelty Delta Analysis for Reviewer Support - Part 1}
\label{fig:novelty-prompt-1}
\end{figure*}

\begin{figure*}[t]
\centering
\begin{tcolorbox}[
    enhanced,
    width=\textwidth,
    colback=white,
    colframe=black,
    boxrule=0.8pt,
    arc=4pt,
    title=\textbf{Novelty Delta Analysis for Reviewer Support - Part 2},
    coltitle=white,
    colbacktitle=black,
    toptitle=1mm,
    bottomtitle=1mm,
    fonttitle=\bfseries,
    breakable
]
\lstset{
    basicstyle=\ttfamily\small,
    breaklines=true,
    frame=none,
    columns=fullflexible,
    keepspaces=true
}
\begin{lstlisting}
5. CRITICAL ASSESSMENT CONSIDERATIONS
   - Identify aspects where claimed novelty may be overstated
   - Analyze whether authors' characterizations of their own novelty align with evidence
   - Consider whether empirical improvements might result from factors other than claimed 
     innovations
   - Assess whether terminology differences might mask conceptual similarities
   - Identify instances where "extensions" might be routine adaptations
   - Note: Frame these as considerations rather than definitive judgments

6. RELATED WORK CONSIDERATIONS
   - Identify potentially relevant work not addressed in the submission
   - Highlight areas where additional comparisons are necessary
   - Note incomplete or potentially misleading characterizations of prior work
   - Identify when claimed "limitations" of prior work may be exaggerated
   - Compare how authors cite specific works versus how they actually relate
   - Note: Present these as information that might help complete the picture

7. KEY OBSERVATION SUMMARY
   - Highlight the most significant independently verified differences from prior work
   - Summarize the main relationships to existing research
   - Identify which claimed contributions have the strongest and weakest differentiation
   - Note the most important discrepancies between author characterizations and 
     independent assessment
   - Note: Frame as observations to inform the reviewer's independent judgment

## Evidence Standards
For each observation, provide:
- Specific references to prior work
- Clear distinction between author claims and independently verified differences
- Explicit identification of similarities and differences based on technical details
- Assessment of whether differences appear substantive or superficial
- Analysis of accuracy in how authors characterize related work

## Example Format for Citation Analysis
"For [Paper X], the authors characterize it as 'limited to simple datasets' and claim 
their work 'extends X to complex scenarios.' The citation sentences appear in the 
following contexts:
- 'Unlike X, which only works on simple datasets, our approach handles complex 
  scenarios' (Introduction)
- 'X proposed the basic framework, but did not address challenge Y' (Related Work)

Independent analysis suggests that Paper X actually did address complex scenarios 
in Section 3.2, though using different terminology. The authors' characterization 
appears to understate X's capabilities to emphasize their contribution. The actual 
primary difference appears to be [specific technical difference] rather than the 
complexity of supported scenarios."

Remember that your role is to provide objective analysis that helps reviewers make 
informed judgments about novelty. Carefully examine both what authors explicitly 
claim and how they implicitly position their work through their characterizations 
of prior research.

{structured_representation}

## Papers from related work not cited
{not_cited_paper_titles}

##Citation Context
{citation_contexts}

## Research Landscape
{research_landscape}
\end{lstlisting}
\end{tcolorbox}
\caption{Novelty Delta Analysis for Reviewer Support - Part 2}
\label{fig:novelty-prompt-2}
\end{figure*}

\begin{figure*}[t]
\begin{tcolorbox}[
   enhanced,
   width=\textwidth,
   colback=white,
   colframe=black,
   boxrule=0.8pt,
   arc=4pt,
   title=\textbf{Reviewer Summary Prompt},
   coltitle=white,
   colbacktitle=black,
   toptitle=1mm,
   bottomtitle=1mm,
   fonttitle=\bfseries,
]
\lstset{
   basicstyle=\ttfamily\small,
   breaklines=true,
   frame=none,
   columns=fullflexible,
   keepspaces=true
}
\begin{lstlisting}
Summarize the following assessment in 5 sentences for a reviewer reviewing at an AI conference.

## Delta Assessment
{novelty_assessment}
\end{lstlisting}
\end{tcolorbox}
\caption{Reviewer Summary Prompt}
\label{fig:summary-prompt}
\end{figure*}

\begin{figure*}[t]
\centering
\begin{tcolorbox}[
    enhanced,
    width=\textwidth,
    colback=white,
    colframe=black,
    boxrule=0.8pt,
    arc=4pt,
    title=\textbf{Novelty Assessment Normalization Prompt},
    coltitle=white,
    colbacktitle=black,
    toptitle=1mm,
    bottomtitle=1mm,
    fonttitle=\bfseries,
    breakable
]
\lstset{
    basicstyle=\ttfamily\small,
    breaklines=true,
    frame=none,
    columns=fullflexible,
    keepspaces=true
}
\begin{lstlisting}
I'll provide you with a novelty assessment extracted from an academic peer review, 
along with the full review for context. Please reformat the novelty assessment into 
a standardized paragraph that begins with a brief description of the paper's contribution 
before analyzing its novelty.

Example of desired format:
"This paper presents a method for neural network compression using knowledge distillation 
with a focus on mobile applications. The approach has limited novelty, as it largely 
builds upon existing techniques in the literature. While the authors claim their technique 
is the first to combine layerwise distillation with quantization-aware training, similar 
combinations have been explored in prior work by Smith et al. (2022) and Jones et al. (2023). 
The main contribution appears to be a specific implementation detail in how gradient flows 
are managed during the distillation process, but this incremental advance does not 
significantly push the boundaries of the field. The paper would benefit from more clearly 
articulating the specific differences from existing approaches to better establish its contribution."

Full review (for context):
{full_review}

Extracted novelty assessment to be reformatted:
{novelty_statements}

Important guidelines:
1. Begin with a clear description of what the paper presents/proposes (drawn from the full review if needed)
2. Create a cohesive paragraph that flows from describing the contribution to analyzing its novelty
3. Maintain all novelty claims and critiques from the original assessment
4. Preserve references to prior work and comparisons
5. Keep the reviewer's judgment of novelty level
6. Incorporate relevant context from the full review to provide a complete picture of the novelty assessment
7. Follow the structure of the example paragraph: description first, then novelty analysis
8. Preserve all critical analysis regarding limitations or strengths of novelty claims

Provide the reformatted novelty assessment:
\end{lstlisting}
\end{tcolorbox}
\caption{Novelty Assessment Normalization Prompt}
\label{fig:normalization-prompt}
\end{figure*}

\begin{figure*}[t]
\centering
\begin{tcolorbox}[
    enhanced,
    width=\textwidth,
    colback=white,
    colframe=black,
    boxrule=0.8pt,
    arc=4pt,
    title=\textbf{Core Novelty Judgment Extraction Prompt},
    coltitle=white,
    colbacktitle=black,
    toptitle=1mm,
    bottomtitle=1mm,
    fonttitle=\bfseries,
    breakable
]
\lstset{
    basicstyle=\ttfamily\small,
    breaklines=true,
    frame=none,
    columns=fullflexible,
    keepspaces=true
}
\begin{lstlisting}
Extract 2-3 core novelty judgments from this assessment:

{reference_assessment}

Focus on statements that directly assess:
- How novel/original the contribution is
- How work relates to prior research
- Specific novelty limitations
- Whether advance is incremental/fundamental

Exclude general recommendations or writing suggestions.

For each judgment, explain why it's considered a core novelty assessment.
Provide rationale for your selection of these specific judgments.
\end{lstlisting}
\end{tcolorbox}
\caption{Core Novelty Judgment Extraction Prompt}
\label{fig:prompt-extraction-core-judgments}
\end{figure*}

\begin{figure*}[t]
\centering
\begin{tcolorbox}[
    enhanced,
    width=\textwidth,
    colback=white,
    colframe=black,
    boxrule=0.8pt,
    arc=4pt,
    title=\textbf{Reviewer Novelty Evaluation Prompt},
    coltitle=white,
    colbacktitle=black,
    toptitle=1mm,
    bottomtitle=1mm,
    fonttitle=\bfseries,
    breakable
]
\lstset{
    basicstyle=\ttfamily\small,
    breaklines=true,
    frame=none,
    columns=fullflexible,
    keepspaces=true
}
\begin{lstlisting}
Compare reviewer assessment against reference using these core judgments:

Core Judgments: {extracted_core_judgments}
Reference: {reference_assessment}
Reviewer: {reviewer_assessment}

Evaluate three dimensions:

1. JUDGMENT SIMILARITY: Do they identify same novelty strengths/weaknesses?
   - For each core judgment, find corresponding judgment in reviewer assessment
   - Assess similarity and provide detailed explanation of alignment/differences
   - Include confidence score for each comparison
   - If the core judgement is referring to a very specific aspect of the methodology 
     and the reviewer assessment does not mention it, then the core judgment is 
     not similar to the reviewer assessment.

2. CONCLUSION ALIGNMENT: Same bottom-line about novelty sufficiency?
   - Determine overall conclusions (SUFFICIENT / INSUFFICIENT / MIXED)
   - Explain whether conclusions align and why

3. PRIOR_WORK_ENGAGEMENT:
   - How does the reviewer engage with prior work?
   - Does the reviewer mention prior work?
   - Does the reviewer compare the current work to prior work?
   - Does the reviewer provide evidence for their claims?
   - Does the reviewer use prior work to support or critique the work?
   - Evaluate number and relevance of citations to prior work 
     (NONE: no citations; LIMITED: 1 to 2; EXTENSIVE: 3+ relevant citations).

4. DEPTH_OF_ANALYSIS:
   - Assesses how deeply specific novelty aspects are compared to prior work
     (SURFACE LEVEL: vague; MODERATE: 1 to 2 aspects; DEEP: 3+ or highly detailed comparisons)

Provide explanations for all assessments to support reasoning.
\end{lstlisting}
\end{tcolorbox}
\caption{Reviewer Novelty Evaluation Prompt}
\label{fig:prompt-reviewer-evaluation}
\end{figure*}